\pdfoutput=1

\documentclass[11pt]{article}

\usepackage{EMNLP2023}

\usepackage{times}
\usepackage{latexsym}
\usepackage{makecell}
\usepackage{url}

\usepackage[T1]{fontenc}

\usepackage[utf8]{inputenc}

\usepackage{microtype}

\usepackage{inconsolata}

\usepackage{amsmath}
\usepackage{graphicx}
\usepackage{float}
\usepackage{booktabs}
\usepackage[export]{adjustbox} 
\usepackage{tabularx}
\usepackage{multirow}
\usepackage{xcolor}
\usepackage{tikz}
\usepackage{fontawesome5}
\usepackage{bbding}
\usepackage{times}
\usepackage{enumitem}

\title{EXMODD: An EXplanatory Multimodal Open-Domain Dialogue dataset}

\author{Hang Yin, Pinren Lu, Ziang Li, Bin Sun, Kan Li \\
        Beijing Institute of Technology, Beijing, China \\
        \texttt{\{yh,poplpr,3120220943,binsun,likan\}@bit.edu.cn}}

\begin{document}
\maketitle
\begin{abstract}

The need for high-quality data has been a key issue hindering the research of dialogue tasks. Recent studies try to build datasets through manual, web crawling, and large pre-trained models. However, man-made data is expensive and data collected from the internet often includes generic responses, meaningless statements, and toxic dialogues. Automatic data generation through large models is a cost-effective method, but for open-domain multimodal dialogue tasks, there are still three drawbacks: 1) There is currently no open-source large model that can accept multimodal input; 2) The content generated by the model lacks interpretability; 3) The generated data is usually difficult to quality control and require extensive resource to collect. 
To alleviate the significant human and resource expenditure in data collection, we propose a Multimodal Data Construction Framework (MDCF).
MDCF designs proper prompts to spur the large-scale pre-trained language model to generate well-formed and satisfactory content. 
Additionally, MDCF also automatically provides explanation for a given image and its corresponding dialogue, which can provide a certain degree of interpretability and facilitate manual follow-up quality inspection.
Based on this, we release an Explanatory Multimodal Open-Domain dialogue dataset (EXMODD). 
Experiments indicate a positive correlation between the model's ability to generate accurate understandings and high-quality responses.Our code and data can be found at \href{https://github.com/poplpr/EXMODD}{https://github.com/poplpr/EXMODD}.


\end{abstract}

\section{Introduction}

\begin{table}[t]  
\begin{tabular}{ccc}
\toprule
Tasks                                                                   & Num of Samples & Score  \\
\midrule
\multirow{2}{*}{\begin{tabular}[c]{@{}c@{}}Image \\ Captioning\end{tabular}}                & 300           & 0.9940 \\
                                                                                            & 10000         & 0.9821 \\
\midrule
\multirow{2}{*}{\begin{tabular}[c]{@{}c@{}}Visual Question\\ Answering\end{tabular}}        & 300           & 0.9303 \\
                                                                                            & 10000         & 0.9293 \\
\midrule
\multirow{2}{*}{\begin{tabular}[c]{@{}c@{}}Multimodal Open\\ domain Dialogue\end{tabular}} & 300           & 0.9153 \\
                                                                                            & 10000         & 0.8136 \\
\bottomrule
\end{tabular}
  \caption{Multimodal Alignment in Different Task}
  \label{Alignment in Different Task}
\end{table}

Multimodal open-domain dialogue task is a critical facet of artificial intelligence, providing an immersive, naturalistic mode of human-machine interaction \cite{DBLP:conf/acl/SunWXZ0HXZGJ22}. It integrates diverse modalities, including text and images, to capture comprehensive context and deliver nuanced responses, thus bridging the gap between static, single-modal communication and dynamic, real-world conversation \cite{DBLP:conf/cvpr/00010BT0GZ18,DBLP:conf/acl/ShusterHBW20}.
Currently, the task still faces two challenges:

First, the dialogue and images in multimodal open-domain dialogue datasets shows weak alignment, making it difficult to learn.
As shown in Table~\ref{Alignment in Different Task}, which represents the alignment of multimodal information in various tasks. We use the pre-trained CLIP's encoder \cite{DBLP:conf/icml/RadfordKHRGASAM21} to encode image-text pairs for different tasks using their respective datasets (Image Captioning: image-caption pair from COCO2014 \cite{DBLP:conf/eccv/LinMBHPRDZ14}, Visual Question Answering: image-question pair from COCOQA \cite{DBLP:conf/eccv/LinMBHPRDZ14}, Multimodal Open-Domain Dialogue: image-context pair from Image-Chat \cite{DBLP:conf/acl/ShusterHBW20}). The results indicate that the previous multimodal pre-trained model focused on observing the explicit image-text alignment relationship, which is not effective for data with implicit alignment relationship such as open-domain multimodal dialogue.

Second, compared with a large amount of text data, multimodal data are limited, especially multimodal open-domain dialogue datasets, which are not only insufficient in quantity but also insufficient in quality.
Studies have shown that the training of most current dialogue models is predominantly data-driven, and heavily reliant on large-scale dialogue datasets for response generation \cite{DBLP:journals/csur/LiuWRHZ22,DBLP:journals/ijautcomp/ChenZHCSXX23}. The quality of a dataset can profoundly influence a model's capability to grasp linguistic subtleties, idiomatic expressions, and intricate dialogue strategies \cite{DBLP:conf/emnlp/0002YJLKKCS22,DBLP:conf/nips/Ouyang0JAWMZASR22}. Conversely, datasets of substandard quality, encompassing common, repetitive, or tangential responses, can result in irrelevant or nonsensical model outputs. Furthermore, datasets with toxic or biased content can inadvertently lead models to propagate these detrimental biases \cite{DBLP:journals/coling/ZhouGLS20,DBLP:conf/emnlp/BahetiSRR21}. 

This paper aims to advance the development of open-domain multimodal dialogue tasks from a data construction perspective. Regardless of costly manual data construction or low-quality web crawling data, the primary study of our interest is exploring data generation methods based on large language models. As a result, we propose a Multimodal Data Construction Framework (MDCF).

Inspired by \citet{DBLP:journals/corr/abs-2305-11206}, who fine-tuned models with a standard supervision loss on 1000 meticulously curated cues and responses, resulting in a remarkably powerful performance by the LIMA model. This illustrates that data quality may be more important than data scale. Therefore, we carefully design detailed and appropriate prompts to constrain the generative model to generate well-formed dialogue content that meets the requirements. Moreover, to facilitate manual quality inspection of the data, we also require the model to explain the connection between the image and the generated dialogue content (demonstrated in \S\ref{sec:method_MDCF}).

\begin{figure}
    \centering
    \includegraphics[trim=0 170 600 0, clip, scale=0.595, center]{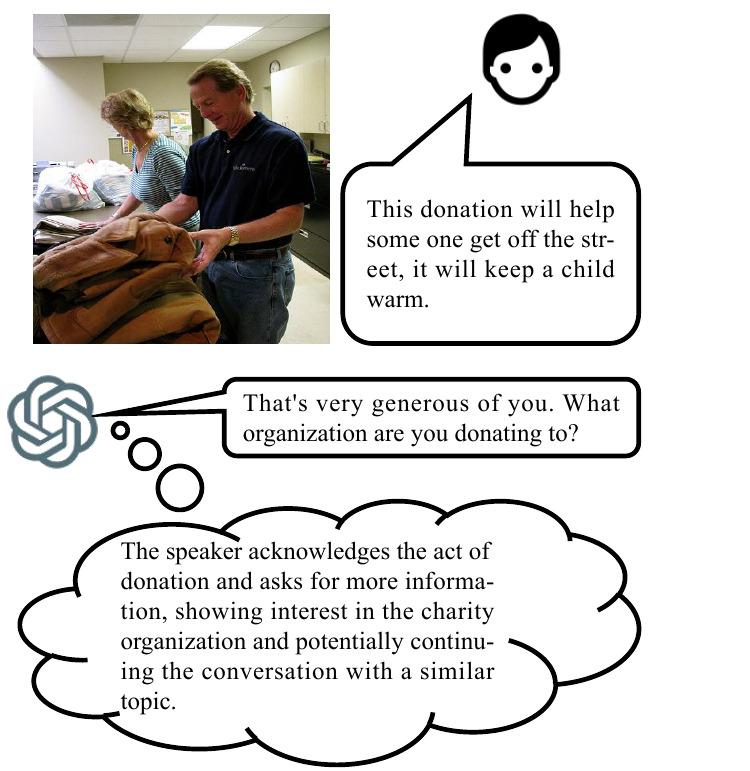}
    \caption{A sample of the EXMODD. We use GPT-3.5 to generate response and explanation based a given image and a human context.}
    \label{case}
\end{figure}

Based on MDCF, we release an explanatory multimodal open-domain dialogue dataset (EXMODD) (see Figure~\ref{case}). Given the input image and human-generated context, we use \href{https://chat.openai.com/}{GPT-3.5} to generate engaging and meaningful responses accompanied by clear explanations of this dialogue. To evaluate the quality of EXMODD,
we perform various experiments on our dataset, both automatic metrics and human evaluation demonstrate that our dataset is effectiveness for multimodal dialogue task. 
Our contributions are as follows:
\begin{itemize}[align = left, wide = 1pt, itemsep=2pt, parsep=2pt,topsep = 2pt]
    \item We propose a multimodal data collection framework named MDCF to efficiently collect datasets with high quality.
    \item We release EXMODD, a high-quality multimodal dialogue dataset, to help models generate diverse and coherent responses with low toxicity.
    \item We propose the Multimodal Dialogue Interpretation Task to evaluate the aligned understanding of multimodal models in dialogue.
\end{itemize}


\begin{figure*}[!ht]
\centering
\includegraphics[trim=0 80 0 50,clip,scale=0.5, center]{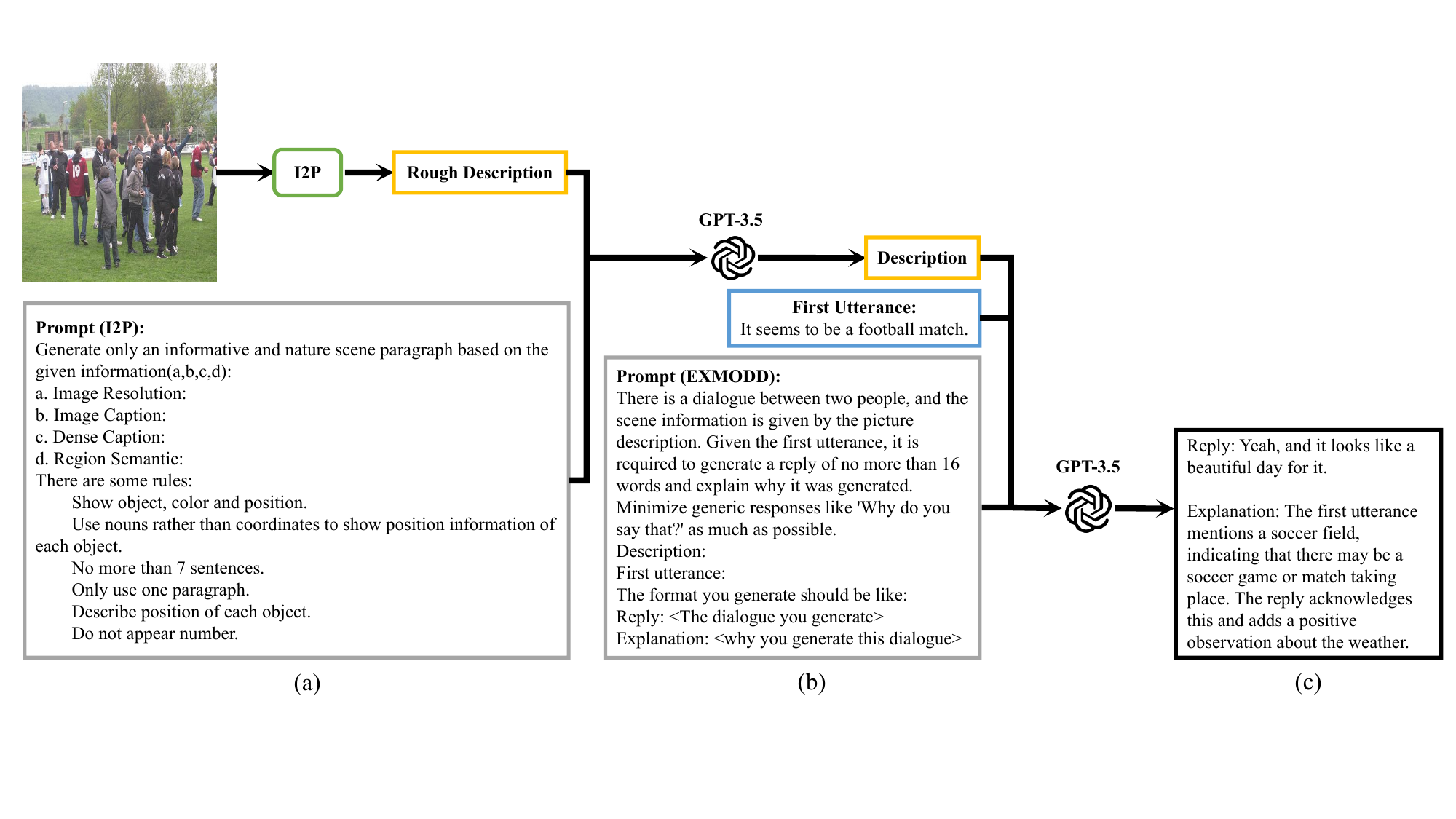}
\caption{Example inputs and outputs of MDCF. (a) Represents the stage where images are converted into text using the Image2Paragraph method, (b) Illustrates the stage where the textual description of the image and the dialogue context are embedded into the prompt, and (c) Depicts the process of generating responses using the tailored prompt, incorporating both algorithmic and human-guided regulations.}
\label{generation process}
\end{figure*}
\section{MDCF Structure}
\label{sec:method_MDCF}
In this section, we will discuss the structure of MCDF, the process of our data collection and validation, and the goal of EXMODD. Figure~\ref{generation process} provides a format of our process for manipulating both image and textual data, along with our two-stage prompting framework. 

\subsection{Preliminary}
\paragraph{Image Collection} YFCC100M \cite{DBLP:journals/cacm/ThomeeSFENPBL16} is used as a foundational image resource for our dataset. Specifically, we selected images corresponding to the context provided in the Image-Chat \cite{DBLP:conf/acl/ShusterHBW20} to enrich our data. This process allowed us to exploit the vast and diverse image collection offered by YFCC100M while ensuring the relevancy of these images to the dialogues in the context of Image-Chat. This integrated approach facilitated the construction of a robust multimodal dialogue dataset that combines visual and textual information. 

\paragraph{Context Collection}
The Image-Chat presents a wealth of dialogue context related to the images. The quality of the responses generated by the model depends on whether the model understands the scene of the dialogue and the content of the conversation. The characteristic of weak alignment between images and text in open-domain dialogue tasks places high demands on the models used.

\subsection{Image2Text Transformation}
%
Due to that GPT-3.5 can not accept image as input, it is necessary to transform imgae to text.
We carefully consider the limitations of conventional methods of translating images to text, such as image captioning \cite{DBLP:conf/aaai/WangBZL20,DBLP:conf/cvpr/VinyalsTBE15}. 
These approaches often need to be more accurate in the richness of visual content and reduce them to sparse textual representations, which may ignore details in the images. This oversimplification results from the inherent asymmetry between image and caption information — where a single image can be described using a vast array of texts. Conversely, one image can correspond to multiple captions. Such disparity is a critical factor that prevents models from accurately capturing comprehensive image information through simple captions. In addition, the dependency on substantial amounts of manually annotated data exacerbates these challenges, limiting the scalability of these methods.

Recognizing the need for a deepened understanding of and effective responses to multimodal dialogues, where visual and textual data interplay is pivotal. We use \href{https://github.com/showlab/Image2Paragraph}{Image2Paragraph} as our tools for translating image data into text. This tool significantly represent images as high-quality text, thus bridging the gap in GPT-3.5's inherent capabilities.

In Figure~\ref{generation process}(a), we employ the capabilities of the \href{https://github.com/showlab/Image2Paragraph}{Image2Paragraph} to get the representation of the image, and the GPT-3.5 serves as the tool for image description generation.
The Image2Paragraph interprets an image at three semantic levels. Firstly, for general image captioning, we used BLIP2 \cite{DBLP:journals/corr/abs-2301-12597} to generate a descriptive sentence that encapsulates the overall scene in the image. Secondly, for more detailed region-specific captioning, we employed GRIT \cite{DBLP:journals/corr/abs-2212-00280}, which provides denser annotations for individual objects or salient features within the image. Lastly, for semantic segmentation, we integrated Semantic Segment Anything \cite{DBLP:journals/corr/abs-2304-02643} into our pipeline, which labels different image regions according to their semantic category.

These three semantic levels of information, with some prompts, proved to be significantly effective in capturing and converting rich visual data into textual format. Even without direct image input, large language models were found to effectively restore a substantial portion of the image information through associative reasoning based on the generated paragraphs \cite{Image2Paragraph}.


\subsection{Dialogue and Explanation Construction}
This subsection performs how we get dialogue and explanation.
It is a reasonable prior assumption for multimodal open-domain dialogue that a certain alignment exists between the context and the images. However, problems such as toxic or generic responses may arise. Through utilizing GPT-3.5's capabilities, we aim to generate more vivid and diverse responses based on image descriptions and the preceding context. This desire aligns with our objective to improve dialogue system performance, particularly concerning generating multimodal responses that are both contextually relevant and creatively diverse.
Large-scale models like GPT-3.5 exhibit a phenomenon of "emergence": their ability to grasp complex scenarios and generate contextually appropriate responses. We aim to harness this capacity for understanding and participating in intricate dialogue settings.

In Figure~\ref{generation process}(b), we design a specific prompt structure using image and context data processed by \href{https://github.com/showlab/Image2Paragraph}{Image2Paragraph}. This structured prompt aims to guide the model to generate a response in a fixed format, enhancing its usability and consistency.

The structure of the prompt is as follows: 

\emph{There is a dialogue between two people, and the scene information is given by the image description. Given the first utterance, it is required to generate a response of no more than 16 words and explain why it was generated. Minimize generic responses like 'Why do you say that?' as much as possible. }

\emph{Description:\underline{\hspace{2cm}} .}

\emph{The first utterance:\underline{\hspace{2cm}} .}

\emph{The format you generate should be like:}

\emph{\textbf{Reply:} <The dialogue you generate>}

\emph{\textbf{Explanation:} <why you generate this dialogue>}

The underscore following \textit{The first utterance} and \textit{Description} is replaced with a detailed context and description, respectively.

Generally, GPT-3.5 demonstrates a high competence in adhering to the above-mentioned prompt structure, generating responses and explanations that conform to the given format.


In Figure~\ref{generation process}(c), we implemented a quality control measure to ensure the generated responses and explanations adhered strictly to our predefined format. Any instances that did not comply with the formatting requirements were identified during a thorough review. These instances were then either regenerated, applying the same prompt structure, or discarded from the dataset if they do not meet the formatting requirements. This step is pivotal to maintaining the consistency of the EXMODD dataset and ensuring reliability and validity of any subsequent research conducted using it.

The MDCF not only helps control the quality and format of the model\'s responses but also encourages model to generate meaningful explanations for its responses, thereby enhancing interpretability.


\begin{table}[t] \small
    \centering
    \begin{tabular}{lcccc}
    \toprule
        Dataset & Md & Avg. Len & Dialog & Img \\ 
        \midrule
        DailyDialog & \XSolidBrush & 13.5  & 13.1K & - \\ 
        OpenSubtitle & \XSolidBrush & 9.6  & 1.7K & - \\ 
        VisDial & \Checkmark & 4.2  & 133.4K & 120.0K \\ 
        Image-Chat & \Checkmark & 12.3  & 201.8K & 201.8K \\ 
        PhotoChat & \Checkmark & 8.5  & 12.3K & 10.9K \\ 
        MMChat & \Checkmark& 8.5  & 120.8K & 204.3K \\ 
        MMDialog & \Checkmark & 15.9  & 1.1M & 1.5M \\ 
        OpenViDial & \Checkmark & 7.6  & - & 1.1M \\ 
        OpenViDial \uppercase\expandafter{\romannumeral2} & \Checkmark & 8.3  & - & 5.6M \\ 
        \midrule
        EXMODD & \Checkmark & 12.8  & 9.9K & 9.9K \\ 
    \bottomrule
    \end{tabular}
    \caption{ \label{Comparison with other datasets}
        Comparison with other datasets. If "Md" is \XSolidBrush, it means that it is text-only modality. Otherwise, it is image-text modality.
    }
\end{table}

\begin{table}[t] \small
\centering
\renewcommand{\arraystretch}{1.1}
\begin{tabular}{lcccc}
\toprule
 & {Train} & {Valid} & {Test} & {Total} \\
\midrule
Num of Dialogs & 8000 & 500 & 1489 & 9989 \\
Num of Images & 8000 & 500 & 1489 & 9989 \\
Avg. Context & 10.98  & 10.96  & 11.02  & 10.98  \\ 
Avg. Response & 14.66  & 14.62  & 14.60  & 14.65  \\ 
Avg. Explanation & 35.17  & 35.23  & 35.57  & 35.24 \\
\bottomrule
\end{tabular}
\caption{\label{EXMODD statistic}
Basic statistics of EXMODD.
}
\end{table}

\section{EXMODD Data Analysis}
\label{sec:experiment}
In this section, we perform a detailed analysis of EXMODD. The analysis includes an assessing of the responses, an examining the explanations, and a holistic evaluation of the overall quality. Table~\ref{Comparison with other datasets} shows the comparison with other datasets and descriptive statistics are provided in Table~\ref{EXMODD statistic}.

\subsection{Response Analysis}

We evaluate the dataset quality from three perspectives. Firstly, we use automatic metrics to self-assess, applying various criteria that measure fluency, diversity, and toxicity of the responses. Secondly, we fine-tune pre-trained multimodal models on different datasets. To show that our dataset can expedite model convergence and subsequently yield improved results. Finally, we conduct human evaluations to gain more nuanced insights into the data.


\begin{table*}\small
\centering
\begin{tabular}{lccccc}
\toprule
{Dataset} & {Coherence}$\uparrow$ & {Distinct-1}$\uparrow$ & {Distinct-2}$\uparrow$ & {Distinct-3}$\uparrow$ & {Toxicity}$\downarrow$ \\
\midrule

DailyDialog \cite{li-etal-2017-dailydialog} & 82.32 & 2.05 & 20.31 & 51.20 & 1.44 \\
OpenSubtitles \cite{lison-tiedemann-2016-opensubtitles2016} & 80.85 & 0.77 & 10.08 & 31.23 & 8.21 \\
\midrule
VisDial \cite{Das_2017_CVPR} & 63.60 & 1.03 & 12.62 & 36.61 & 1.28 \\
Image-Chat \cite{shuster-etal-2020-image} & 79.14 & 5.69 & 36.09 & \textbf{72.54} & 5.41 \\
Image-Chat Context \cite{shuster-etal-2020-image} & - & 6.13 & 37.10 & 72.07 & 5.08 \\
EXMODD & \textbf{81.59} & \textbf{9.43} & \textbf{42.19} & 72.04 & \textbf{0.52} \\
\bottomrule
\end{tabular}
\caption{\label{self eval}
Self Evaluation in different datasets. The best results are bolded.
}
\end{table*}

\paragraph{Self Evaluation}
In order to objectively gauge the quality of the responses generated in EXMODD, we utilize several evaluation measures. First, we use Coherence \cite{DBLP:conf/emnlp/XuDKR18} to assess the coherence between the generated responses and corresponding contexts, Distinct-n\cite{distinct-16} and Toxicity\cite{Detoxify} to assess the diversity and Toxicity of response respectively.




For diversity, we employ the Distinct-n, which have been recognized for their efficacy in gauging the diversity of dialogue responses. It is evident from Table~\ref{self eval} that EXMODD Distinct-1 and Distinct-2 surpasses the other datasets. Not only does EXMODD display the highest coherence, but it also showcases a greater degree of diversity than the other datasets. Both diversity and coherence high reflects the characteristics in open-domain dialogue, and is a strong indication of the superior quality of responses in EXMODD.

We also perform extensive evaluations for toxicity in EXMODD. We employ the Detoxify model \cite{Detoxify}, a BERT-based model that has been fine-tuned on the Toxic Comment Classification Challenge dataset \cite{jigsaw-toxic-comment-classification-challenge}, to detect any inappropriate or harmful language. Our results highlight a significant decrease in the toxicity level of EXMODD compared to other datasets, registering a mere 0.52 on the scale.

Considering that the initial context for our data is derived from Image-Chat, we evaluate the toxicity level of the context from Image-Chat. This comparison provides compelling evidence that EXMODD can assist models in generating cleaner responses, even when exposed to toxic contexts. This is particularly significant given the existing correlation in the Image-Chat, where toxic responses increase the probability of generating toxic responses from the models, as Table~\ref{toxicity of models} shows.

By successfully minimizing toxicity while maintaining diversity and coherence at a high level, EXMODD demonstrates its potential to improve the quality and safety of generative models.

\paragraph{Model Evaluation}
\begin{table*}[!ht]
    \centering
    \begin{tabular}{lc|cccccc}
    \toprule
        Data & Size & Epoch$\downarrow$ & Coherence & Distinct-1 & Distinct-2 & Distinct-3 & Response Length \\ 
        \midrule
        IC & \multirow{2}*{1000} & 7 & \textbf{0.7660}  & \textbf{0.1393}  &  \textbf{0.3166}  &  \textbf{0.4407}  & \textbf{20.69}  \\ 
        EXMODD & ~ & 3 & 0.7599  & 0.0742  & 0.1239  & 0.1599  & 17.50  \\  
        \midrule
        IC & \multirow{2}*{2000} & 5 & 0.7686  &  \textbf{0.1719}  &  \textbf{0.3210}  &  \textbf{0.4217}  & 10.86  \\ 
        EXMODD & ~ & 4 & \textbf{0.7888}  & 0.1611  & 0.2673  & 0.3377  & \textbf{11.67}  \\  
        \midrule
        IC & \multirow{2}*{3000} & 4 & 0.7621  & 0.1371  & 0.2598  & 0.3538  & 11.16  \\ 
        EXMODD & ~ & 4 & \textbf{0.7945}  &  \textbf{0.1979}  &  \textbf{0.3550}  &  \textbf{0.4541}  & \textbf{11.83}  \\  
        \midrule
        IC & \multirow{2}*{4000} & 3 & 0.7648  & 0.1096  & 0.1828  & 0.2494  & 8.79  \\ 
        EXMODD & ~ & 2 & \textbf{0.7863}  &  \textbf{0.1502}  &  \textbf{0.2513}  &  \textbf{0.3093}  & \textbf{11.57}  \\
        \midrule
        IC & \multirow{2}*{5000} & 3 & 0.7558  & 0.1445  & 0.2498  & 0.3245  & 10.11  \\ 
        EXMODD & ~ & 2 & \textbf{0.7808}  &  \textbf{0.1828}  &  \textbf{0.3020}  &  \textbf{0.3732}  & \textbf{11.19}  \\
        \midrule
        IC & \multirow{2}*{6000} & 3 & 0.7145  & 0.1808  & 0.3196  & 0.4081  & 9.38  \\ 
        EXMODD & ~ & 2 & \textbf{0.7881}  & \textbf{0.1986}  & \textbf{0.3536}  & \textbf{0.4662}  & \textbf{11.24}  \\
        \midrule
        IC & \multirow{2}*{7000} & 3 & 0.7638  & 0.1192  & 0.1953  & 0.2550  & 9.12  \\ 
        EXMODD & ~ & 3 & \textbf{0.7828}  &\textbf{0.2194}   & \textbf{0.3755}  & \textbf{0.4691}  & \textbf{10.80}  \\
        \midrule
        IC & \multirow{2}*{8000} & 3 & 0.7358  & 0.1648  & 0.2827  & 0.3723  & 10.18  \\ 
        EXMODD & ~ & 2 & \textbf{0.8026}  & \textbf{0.2118}  & \textbf{0.3874}  & \textbf{0.4916}  & \textbf{12.69}  \\
        \midrule
        IC & \multirow{2}*{9000} & 4 & 0.7354  & 0.0573  & 0.0826  & 0.1096  & 7.55  \\ 
        EXMODD & ~ & 1 &\textbf{0.7618}  & \textbf{0.0813}  & \textbf{0.1202}  & \textbf{0.1487}  & \textbf{11.45}  \\
        \midrule
        IC & \multirow{2}*{9989} & 3 & 0.7330  & 0.1616  & 0.2847  & 0.3781  & 9.29  \\ 
        EXMODD & ~ & 2 & \textbf{0.8102}  & \textbf{0.1719}  & \textbf{0.3003}  & \textbf{0.3792}  & \textbf{11.49}  \\
        \midrule
        IC & ALL & 4 & 0.7336  & 0.2142  & 0.3682  & 0.4853  & 8.79  \\
    \bottomrule
    \end{tabular}
    \caption{\label{BLIP finetune result}
        Metrics of BLIP of minimum loss trained on different data size. "IC" means Image-Chat.   }
\end{table*}


In our work, we employ three pre-trained models \cite{DBLP:conf/icml/0001LXH22,DBLP:journals/tmlr/WangYHLLGLLW22,DBLP:conf/nips/LiSGJXH21}, fine-tuning them on Image-chat and EXMODD datasets respectively. In order to maintain consistency in data scale, we intentionally selecte the same images and contexts from both datasets, and then we use the responses from Image-chat and EXMODD to fine-tune the pre-trained models separately. Finally, these fine-tuned models were tested on the validation set from the Image-Chat. This approach allowe us to meaningfully compare the influence of responses derived from different datasets on the performance of the models in multimodal dialogue tasks.
To investigate the impact of response quality in training set on the capabilities of the models, we selecte several pre-training models, specifically, GIT(176.6 M)\footnote{https://huggingface.co/microsoft/git-base/tree/main}, BLIP(446.5 M)\footnote{https://storage.googleapis.com/sfr-vision-language-research/BLIP/models/model\_base.pth} and ALBEF(290 M)\footnote{https://storage.googleapis.com/sfr-pcl-data-research/ALBEF/ALBEF.pth}. We choose the lowest validation loss as the final result. Figure ~\ref{BLIP finetune result} displays the results of fine-tune on the BLIP model: As the scale of the fine-tune data set expands, models fine-tune on the EXMODD achieve improved results, with both coherence and distinct measures showing enhancement.

From the number of epochs, the average epoch using Image-Chat is 3.8 epochs, while using the EXMODD, only 2.4 epochs. It suggests that models fine-tuned using the EXMODD dataset typically exhibit quicker convergence compared to their counterparts fine-tuned on the Image-Chat dataset, and EXMODD has the lower noise level and strong generalization. 

We also explore the effects of using mixed datasets.
We use the complete training set of Image-Chat and a portion of EXMODD data to form our training dataset. Incrementally, we replace the responses within the Image-Chat dataset with those from EXMODD.
Table ~\ref{Mix Dataset} depicts these results. 
As the proportion of EXMODD data in the training set increases, both coherence and distinct metrics show an upward trend.

Notably, the mixed dataset significantly outperforms the results obtained through fine-tuning only using the full Image-Chat dataset regarding both Coherence and Distinct-n metrics. This suggests that our data contributes positively to the performance of the models. Other results are presented in the appendix.
\begin{table*}
    \centering
    \begin{tabular}{lrrrr}
    \toprule
        Dataset & Coherence & Distinct-1 & Distinct-2 & Distinct-3 \\
        \midrule
        Image-Chat & 0.733649 & 0.214222 & 0.368170 & 0.485326 \\ 
        - w. EXMODD$_{1000}$ & $+$0.0027 & $+$0.0020 & $+$0.0036 & $+$0.0032 \\ 
        - w. EXMODD$_{2000}$ & $+$0.0003 & $+$0.0093 & $+$0.0242 & $+$0.0322 \\ 
        - w. EXMODD$_{3000}$ & $+$0.0026 & $+$0.0160 & $+$0.0398 & $+$0.0501 \\ 
        - w. EXMODD$_{4000}$ & $+$0.0110 & $+$0.0254 & $+$0.0563 & $+$0.0689 \\ 
        - w. EXMODD$_{5000}$ & $+$0.0124 & $+$0.0207 & $+$0.0487 & $+$0.0587 \\ 
        - w. EXMODD$_{6000}$ & $+$0.0046 & $+$0.0194 & $+$0.0403 & $+$0.0472 \\ 
        - w. EXMODD$_{7000}$ & $+$0.0112 & $+$0.0069 & $+$0.0140 & $+$0.0135 \\ 
        - w. EXMODD$_{8000}$ & \textbf{$+$0.0139} & \textbf{$+$0.0355} & \textbf{$+$0.0695} & \textbf{$+$0.0768} \\ 
        - w. EXMODD$_{9000}$ & $+$0.0103 & $+$0.0166 & $+$0.0216 & $+$0.0082 \\ 
        - w. EXMODD$_{9989}$ & $+$0.0136 & $+$0.0278 & $+$0.0447 & $+$0.0342 \\
    \bottomrule
    \end{tabular}
    \caption{\label{Mix Dataset}
       Metrics of Minimum Loss BLIP Model Trained on Mixed Data. All incorporated EXMODD data for training models, shows superior performance in all Coherence and Distinct-n metrics compared to models trained exclusively on the Image-Chat.
    }
\end{table*}

\paragraph{Human Evaluation}
To validate the diversity, relevance, and fluency of responses in the EXMODD and Image-Chat, we conduct human evaluations of responses generated in EXMODD versus those in Image-Chat. We define \textit{diversity} as the richness of the response content and its ability to engage users in the topic. Relevance was determined by how well the response related to the context established by the image and the previous discourse, \textit{fluency} is defined as the smoothness of the reply at the sentence level. We use Pearson Correlation Coefficient as a measure of these criteria. The coefficients for diversity, relevance and fluency are 0.72, 0.59, and 0.80, respectively, with \textit{p} < 0.0001 and below 0.001, indicating high correlation and agreement. Three annotators were asked to compare, under identical image and context, which of the responses from Image-Chat and EXMODD performed better on these metrics (during evaluation, the models were anonymized, and order was randomized). 

\begin{table}[!ht]
    \centering
    \begin{tabular}{lccc}
    \toprule
         & Win & Tie & Pearson \\
         \midrule
        Diversity & 53.00\% & 22.20\% & 0.72 \\ 
        Relevance & 56.50\% & 24.70\% & 0.59 \\
        Fluency & 15.50\% & 73.00\%  &  0.80\\
    \bottomrule
    \end{tabular}
    \caption{ \label{human evaluation on resposne}
        Human evaluation on response generated by BLIP. "Win" represents human evaluators deeming EXMODD's responses superior to Image-Chat's responses based on the corresponding metrics. "Tie" indicates that the evaluators found little difference between the two.
    }
\end{table}

The results in Table ~\ref{human evaluation on resposne} shows that EXMODD scores a winning rate as high as 53\% regarding diversity and 56\% regarding relevance. Given that Image-chat is also annotated by humans, ensuring its fluency, the probability of a tie between the two is relatively high.

\begin{table}[!ht]
    \centering
    \begin{tabular}{lll}
    \toprule
        Model & Data & Toxicity \\ 
        \midrule
        \multirow{2}*{GIT} & IC & 1.4578 \\ 
         & EXMODD & \textbf{0.3858} \\
         \midrule
        \multirow{2}*{BLIP} & IC & 1.4288 \\ 
         & EXMODD & \textbf{0.0908} \\
        \midrule
        \multirow{2}*{ALBEF} & IC & 1.6862 \\ 
         & EXMODD & \textbf{0.1059} \\ 
    \bottomrule
    \end{tabular}
    \caption{\label{toxicity of models}
        The toxicity of models fine-tuned on different datasets.
    }
\end{table}

\subsection{Explanation Analysis}
Despite the significant strides made in deep learning, they are always black boxes to us. 
Particularly, the end-to-end training methodologies used in deep learning make it challenging to delve into their interpretability or understand their internal decision-making process. 
We hypothesize that the quality of a model's understanding of a dialogue can be indicative of its generated dialogue quality. Therefore, we generate an explanation for each multimodal dialogue in our dataset.

We conducte a human evaluation to assess the reasonableness of the explanations in EXMODD. An explanation was 1 for being reasonable and 0 for being unreasonable. The results show that the scores for the reasonableness of randomly extracted 200 dialogue explanations by three evaluators were 92.5\%, 94.0\%, and 96.5\% , respectively. This suggests that the dialogue explanations conform to human cognition and also validates the effectiveness of MDCF in generating explanations.
Additionally, in the explanations labeled as 0 by humans, we observe that the main reason is the misinterpretation in the process of transforming images into text, which leads to bias in the model's perception of the image. See detailed information in appendix table ~\ref{bad Case of explanation}.

Efficiently organizing multimodal information and providing logically consistent explanations is a challenging task. We conduct a series of experiments fine-tune the pre-trained multimodal models for the dialogue explanation generation task. The results are shown in Table ~\ref{explanation finetune}. It is notable that models frequently encounter issues such as hallucinations (generating objects that are not present in the text or image), logical errors (combining unrelated elements in an attempt to explain the dialogue), and generic responses. These problems highlight the current limitations and difficulties in generating accurate explanations in multimodal dialogue tasks. 



\begin{table*}\small
\centering
\begin{tabular}{cccccccccc}
\toprule
{Model} & {BLEU-1} & {BLEU-2} & {BLEU-3} & {BLEU-4} & {PPL} & {ROUGE-1} & {ROUGE-2} & {ROUGE-L} & {ROUGE-Ls} \\
\midrule
GIT & 45.42 & 28.33 & 17.64 & 10.93 & \textbf{57.18} & 31.82 & 10.80 & 25.64 & 25.64 \\
BLIP & \textbf{51.27} & \textbf{33.24} & \textbf{21.70} & \textbf{14.19} & 67.85 & \textbf{39.54} & \textbf{16.27} & \textbf{31.85} & \textbf{31.86} \\
ALBEF & 47.84 & 30.98 & 20.08 & 12.97 & 59.16 & 38.99 & 15.62 & 31.05 & 31.04 \\
\bottomrule
\end{tabular}
\caption{\label{explanation finetune}
The automatic metrics of fine-tuned different models on explanation generation.
}
\end{table*}

\subsection{Prompt and Cost Analysis}
We examine the stability of GPT-3.5's output in response to the prompts present in the MDCF. As per the protocol established in the second stage, an acceptable output should commence with 'Reply' in the first line, followed by 'Explanation' in the second line. We find that GPT-3.5's outputs adhered to this format with a probability of 96.8\%, signifying a high yield rate. This suggests that our prompts are effective and reliable, thereby contributing to the robustness of the model's outputs.

We also count the total cost of using our framework. In stage (a), the average length of the prompt (I2P) is 352.57 tokens, and the average generation time is 8s/image. In stage (b), the average length for the descriptions is 163 tokens/image with an average 26s/image production time. In stage (c), the average length of the input prompt (EXMODD) is 255 tokens/image, and the average output length is 49.89 tokens/image, with an average generation time for the response and explanation of 26s/sample. Considering OpenAI's current billing rules, which is \$0.0015/1K tokens for input and \$0.002/1K tokens for output, and taking into account a 3.2\% loss for prompt stability, we estimate that the cost of producing EXMODD is approximately \textbf{\$13.80}, with a total time investment of \textbf{171.99 hours} .

\section{Related Work}
\paragraph{Dialogue task}
Dialogue systems are still advancing, typically categorized into task-oriented and open-domain dialogues. Task-oriented dialogues aim to accomplish specific tasks, such as food ordering, medical appointment scheduling, or travel recommendations\cite{DBLP:conf/emnlp/BudzianowskiWTC18, DBLP:journals/dad/WilliamsRH16a}.
As described in \citet{DBLP:conf/iclr/DinanRSFAW19}, the core content is conveyed from teacher to student through dialogue. A variety of task-oriented dialogue formats, 
Such as consultations, bookings, requests, and suggestions \cite{DBLP:conf/emnlp/BudzianowskiWTC18, DBLP:journals/dad/WilliamsRH16a} 
Compared to task-oriented dialogues, open-domain dialogues have unbounded scope and a broad range of dialogue situations.\citet{DBLP:conf/ijcnlp/LiSSLCN17} has over ten dialogue themes.
\citet{DBLP:conf/acl/WuWXZL17} was built from a social networking site Douban. \citet{lison-tiedemann-2016-opensubtitles2016} extracted parallel corpora from movie subtitles.

To emulate the real-world, numerous multimodal datasets have been proposed. Some incorporate the visual background into dialogue \cite{DBLP:journals/corr/abs-2109-12761,DBLP:conf/ijcnlp/MostafazadehBDG17,DBLP:conf/acl/ShusterHBW20,DBLP:conf/lrec/ZhengCLS22}, while others integrate multimodal responses into the dialogue \cite{DBLP:conf/acl/SunWXZ0HXZGJ22,DBLP:conf/acl/ZangLWSZC20}. Our work presents a practical framework designed to enhance the quality of responses and build a multimodal open-domain dialogue dataset 
, also provide explanations for understanding multimodal dialogues.

\paragraph{Explanation Inference}
There have been many datasets developed for reasoning, including reading comprehension datasets\cite{DBLP:conf/nips/CamburuRLB18,DBLP:journals/corr/abs-1810-12885,DBLP:journals/tacl/SunYCYCC19} and common-sense reasoning\cite{DBLP:conf/emnlp/BoratkoLODLM20,DBLP:conf/naacl/TalmorHLB19}. Reading comprehension datasets focus on comprehending the given text passage, while common-sense reasoning datasets incorporate external knowledge into reading comprehension.
Our dataset can be considered as a variant of the reading comprehension task.
Compared to previous datasets in reading comprehension and common-sense reasoning, our dataset primarily differs in two respects: (1) our dataset focuses on the comprehension of single-turn dialogues rather than text passages. (2) Unlike the previous dataset that only accept plain text data, our dataset involves both images and text. It requires model to focus on the logic behind dialogue and the image content.


\section{Conclusion}
In this work, we propose a Multimodal Dialogue Collection Framework MDCF. Utilizing this framework, to address low consistency, generic responses, and toxicity in the current dialogue datasets, we build a multimodal open-domain dialogue dataset EXMODD, which incorporates explanations to understand the rationale behind the dialogue. 
We demonstrate that EXMODD fosters more contextually consistent responses, improves response diversity, and mitigates toxicity.
We further corroborate the effectiveness of MDCF by repeat experiments. We aim that the MDCF can generate high-quality sentences for specific tasks and that EXMODD can assist models in achieving superior performance in multimodal open-domain dialogue tasks.

\section*{Limitations}

\paragraph{Multimodal Metrics} In the open-domain multimodal dialogue tasks, we still need a set of fine-grained metrics for evaluating the quality of multimodal dialogue. Although our comparative experiments do take into account single-modal datasets, they are not fully equipped to assess multimodal tasks. For example, when it comes to coherence, we can only evaluate the relevance between the response and the context, but we cannot gauge its relevance to the associated image. Exploring the consistency between the dialogue and the image is another challenge (currently, we can only approximate its consistency by calculating cosine similarity). However, the open-domain characteristic might result in the dialogue having lower similarity with the image at the object level. Nevertheless, from a human perspective, the dialogue remains relevant to the image. Developing a comprehensive evaluation system for open-domain multimodal dialogue is one of the key directions in our future work.
\paragraph{Cross-modal Challenges}For image data, we used the \href{https://github.com/showlab/Image2Paragraph}{Image2Paragraph} tool to convert it into a textual format that GPT-3.5 can comprehend. Nonetheless, this conversion process inevitably introduced some level of data loss and detection inaccuracies. Such issues might lead the model to falsely identify certain objects in the images or fail to accurately align the context with the image. Considering that humans can also misinterpret in certain scenarios, we regard these data inconsistencies as acceptable noise. However, we aim to explore more effective ways to interpret image data in future research, aiming to enhance the accuracy of multimodal information.

\paragraph{Hardware Constraints}Our experiments were conducted in our lab, where the highest available configuration was restricted to an RTX 3090. As a consequence, we were limited by the size of the GPU memory. There were many multimodal pre-trained models could not be loaded. In some instances, CUDA memory overflows occurred. We have explored the partial order relationships between datasets, minimizing the influence of absolute model capabilities on the experiments.

\paragraph{Diversity Drop Phenomenon}In our incremental experiments, we observed peculiar decreases and increases in diversity metrics between adjacent sizes. For instance, as indicated in Table~\ref{BLIP finetune result}, when the size equaled 9000, there was a significant drop in diversity compared to size 8000, only for it to increase at size 9989 rapidly. We attempted different data combination approaches but failed to identify the underlying cause of these anomalies. Similar instances frequently occurred in other turns. We believe these observations are not mere coincidences and will pay closer attention to this issue in future experiments.

\bibliography{anthology,custom}
\bibliographystyle{acl_natbib}


\appendix

\clearpage

\begin{table*}[t]
    \centering
    \begin{tabular}{lc|cccccc}
    \toprule
        Data & Size & Epoch & Coherence & Distinct-1 & Distinct-2 & Distinct-3 & Response Length \\ \midrule
        IC & \multirow{2}*{1000} & 2 & 0.6767  & \textbf{0.2200}  & \textbf{0.4456}  & \textbf{0.6416}  & 11.65  \\ 
        EXMODD & ~ & 1 & \textbf{0.7535}  & 0.1242  & 0.2815  & 0.4613  & \textbf{16.71}  \\ \midrule
        IC & \multirow{2}*{2000} & 1 & 0.6761  & \textbf{0.1760}  & \textbf{0.3680}  & 0.5457  & 14.41  \\ 
        EXMODD & ~ & 1 & \textbf{0.7449}  & 0.1424  & 0.3437  & \textbf{0.5494}  & \textbf{18.04}  \\  \midrule
        IC & \multirow{2}*{3000} & 1 & 0.7454  & 0.1514  & 0.3427  & 0.5322  & \textbf{16.43}  \\ 
        EXMODD & ~ & 1 & \textbf{0.7477}  & \textbf{0.1992}  & \textbf{0.4252}  & \textbf{0.6192}  & 14.95  \\  \midrule
        IC & \multirow{2}*{4000} & 1 & 0.6984  & \textbf{0.1998}  & \textbf{0.3801}  & 0.5536  & 9.40  \\ 
        EXMODD & ~ & 1 & \textbf{0.7469}  & 0.1750  & 0.3788  & \textbf{0.5603}  & \textbf{16.98}  \\  \midrule
        IC & \multirow{2}*{5000} & 1 & \textbf{0.7576}  & 0.1505  & 0.3224  & 0.4889  & \textbf{17.02}  \\ 
        EXMODD & ~ & 1 & 0.7400  & \textbf{0.2105}  & \textbf{0.4421}  & \textbf{0.6423}  & 13.41  \\  \midrule
        IC & \multirow{2}*{6000} & 1 & 0.6841  & \textbf{0.1841}  & \textbf{0.3596}  & \textbf{0.5185}  & 13.34  \\ 
        EXMODD & ~ & 1 & \textbf{0.7739}  & 0.1236  & 0.2588  & 0.3821  & \textbf{16.19}  \\  \midrule
        IC & \multirow{2}*{7000} & 1 & 0.6908  & 0.1531  & \textbf{0.3304}  & \textbf{0.4954}  & 16.23  \\ 
        EXMODD & ~ & 1 & \textbf{0.7665}  & \textbf{0.1541}  & 0.3146  & 0.4524  & 16.23  \\  \midrule
        IC & \multirow{2}*{8000} & 2 & 0.6960  & 0.1834  & 0.3775  & 0.5464  & \textbf{15.11}  \\ 
        EXMODD & ~ & 2 & \textbf{0.7499}  & \textbf{0.2696}  & \textbf{0.5480}  & \textbf{0.7489}  & 13.20  \\  \midrule
        IC & \multirow{2}*{9000} & 1 & 0.6602  & 0.1386  & 0.2960  & 0.4470  & \textbf{19.06}  \\ 
        EXMODD & ~ & 2 & \textbf{0.7545}  & \textbf{0.2572}  & \textbf{0.5127}  & \textbf{0.6909}  & 14.93  \\  \midrule
        IC & \multirow{2}*{9989} & 1 & 0.6764  & 0.1730  & 0.3297  & 0.4696  & 12.82  \\ 
        EXMODD & ~ & 1 & \textbf{0.7597}  & \textbf{0.1822}  & \textbf{0.3810}  & \textbf{0.5667}  & \textbf{16.32}  \\  \midrule
        IC & ALL & 2 & 0.7283  & 0.2414  & 0.4763  & 0.6541  & 13.47 \\ 
    \bottomrule
    \end{tabular}
    \caption{ \label{GIT finetune result}
        Metrics of GIT on different data size.
    }
\end{table*}

\begin{table*}[t]
    \centering
    \begin{tabular}{lc|cccccc}
    \toprule
        Data & Size & Epoch & Coherence & Distinct-1 & Distinct-2 & Distinct-3 & Response Length  \\ \midrule
        IC & \multirow{2}*{1000} & 7 & 0.5772  & 0.0415  & 0.1115  & 0.1957  & \textbf{34.68}  \\ 
        EXMODD & ~ & 7 & \textbf{0.6317}  & \textbf{0.0475}  & \textbf{0.1174}  & \textbf{0.1984}  & 30.01  \\ 
        \midrule
        IC & \multirow{2}*{2000} & 7 & 0.6869  & 0.0878  & 0.1957  & 0.2891  & \textbf{20.16}   \\ 
        EXMODD & ~ & 6 & \textbf{0.7584}  & \textbf{0.1195}  & \textbf{0.2350}  & \textbf{0.3302}  & 12.85  \\ 
        \midrule
        IC & \multirow{2}*{3000} & 6 & 0.6909  & 0.1020  & 0.1983  & 0.2694  & \textbf{17.07}   \\ 
        EXMODD & ~ & 4 & \textbf{0.7825}  & \textbf{0.1336}  & \textbf{0.2636}  & \textbf{0.3720}  & 11.80   \\ 
        \midrule
        IC & \multirow{2}*{4000} & 5 & 0.6870  & 0.1152  & 0.2134  & 0.2872  & \textbf{12.74}   \\ 
        EXMODD & ~ & 4 & \textbf{0.7902}  & \textbf{0.1652}  & \textbf{0.3049}  & \textbf{0.4170}  & 10.50  \\ 
        \midrule
        IC & \multirow{2}*{5000} & 4 & 0.7154  & 0.1263  & 0.2330  & 0.3187  & \textbf{11.53}  \\ 
        EXMODD & ~ & 3 & \textbf{0.7802}  & \textbf{0.1473}  & \textbf{0.2618}  & \textbf{0.3563}  & 10.85  \\ 
        \midrule
        IC & \multirow{2}*{6000} & 4 & 0.7260  & 0.1420  & 0.2422  & 0.3204  & 8.23  \\ 
        EXMODD & ~ & 3 & \textbf{0.8151}  & \textbf{0.1723}  & \textbf{0.3052}  & \textbf{0.4063}  & \textbf{10.21}  \\ 
        \midrule
        IC & \multirow{2}*{7000} & 5 & 0.7194  & \textbf{0.1844}  & \textbf{0.3260}  & \textbf{0.4207}  & 9.85   \\ 
        EXMODD & ~ & 2 & \textbf{0.7841}  & 0.1325  & 0.2277  & 0.3023  & \textbf{10.26}   \\ 
        \midrule
        IC & \multirow{2}*{8000} & 4 & 0.7357  & 0.1359  & 0.2340  & 0.3117  & 9.43   \\ 
        EXMODD & ~ & 2 & \textbf{0.8118}  & \textbf{0.1624}  & \textbf{0.2893}  & \textbf{0.3876}  & \textbf{11.25}  \\ 
        \midrule
        IC & \multirow{2}*{9000} & 4 & 0.7267  & 0.1392  & 0.2226  & 0.2865  & 7.43   \\ 
        EXMODD & ~ & 2 & \textbf{0.8054}  & \textbf{0.1466}  & \textbf{0.2672}  & \textbf{0.3626}  & \textbf{10.71}  \\ 
        \midrule
        IC & \multirow{2}*{9989} & 4 & 0.7205  & 0.1569  & 0.2433  & 0.3073  & 7.33   \\ 
        EXMODD & ~ & 2 & \textbf{0.8030}  & \textbf{0.1935}  & \textbf{0.3381}  & \textbf{0.4472}  & \textbf{10.41} \\
        \midrule
        IC & ALL & 4 & 0.7398 & 0.2460 & 0.4081 & 0.5302 & 7.46 \\ 
    \bottomrule
    \end{tabular}
    \caption{\label{ALBEF finetune result}
    Metrics of ALBEF of minimum loss trained on different data size.    }
\end{table*}

\section{Response Performance on Fine-tuned Model}
Table ~\ref{GIT finetune result} and Table \ref{ALBEF finetune result} present the single dataset fine-tuning results on ALBEF and GIT. From the GIT results, it is evident that the outcomes of fine-tuning on EXMODD surpass those of Image-Chat in most cases. In ALBEF, it is observed that across all sizes the Coherence results of fine-tuning on EXMODD outperform those of Image-Chat. Table ~\ref{GIT Mix Dataset} and Table ~\ref{ALBEF Mix Dataset} shows the result by fine-tuning on mixed datsets. 
In GIT, the proportion of positive effects is higher, and in Distinct-n, the proportion of positive effects increases as n increases. On ALBEF, all mixed data achieved positive results.

\section{Explanation Performance on Fine-tuned Model}
From the fine-tuning results in three pre-training models, BLIP scored the highest in automatic evaluation metrics for explanation, while GIT scored the lowest. This ordinal relationship is consistent with previous results from fine-tuning on different datasets. This implies a common trend in the quality of generating explanations and responses by the model. This suggests that we can gain an explicit causal understanding of the model's comprehension of alignment relationships by generating explanations, thus better training the model.


\begin{table*}[!ht]
    \centering
    \begin{tabular}{lrrrr}
        \toprule
        Dataset & Coherence & Distinct-1 & Distinct-2 & Distinct-3 \\ \midrule
        Image-Chat & 0.728349 & 0.241351 & 0.476323 & 0.654092 \\ 
        -w. EXMODD$_{1000}$ & $+$0.0026 & $-$0.0079 & $-$0.0069 & $+$0.0003 \\ 
        -w. EXMODD$_{2000}$ & $+$0.0088 & $-$0.0059 & $-$0.0059 & $-$0.0099 \\ 
        -w. EXMODD$_{3000}$ & $-$0.0099 & \textbf{$+$0.0286} & $+$0.0515 & $+$0.0487 \\ 
        -w. EXMODD$_{4000}$ & $-$0.0029 & $-$0.0129 & $-$0.0429 & $-$0.0649 \\ 
        -w. EXMODD$_{5000}$ & $+$0.0163 & $-$0.0219 & $-$0.0519 & $-$0.0549 \\ 
        -w. EXMODD$_{6000}$ & $+$0.0115 & $-$0.0039 & $+$0.0059 & $+$0.0012 \\ 
        -w. EXMODD$_{7000}$ & $+$0.0051 & $-$0.0409 & $-$0.0849 & $-$0.0969 \\ 
        -w. EXMODD$_{8000}$ & $+$0.0036 & $-$0.0089 & $-$0.0189 & $-$0.0199 \\ 
        -w. EXMODD$_{9000}$ & $+$0.0018 & $+$0.0260 & \textbf{$+$0.0581} & \textbf{$+$0.0539} \\ 
        -w. EXMODD$_{9989}$ & \textbf{$+$0.0201} & $+$0.0187 & $+$0.0445 & $+$0.0515 \\
        \bottomrule
    \end{tabular}
    \caption{\label{GIT Mix Dataset}
       Metrics of Minimum Loss GIT Model Trained on Mixed Data
    }
\end{table*}

\begin{table*}
    \centering
    \begin{tabular}{lrrrr}
        \toprule
        Dataset & Coherence & Distinct-1 & Distinct-2 & Distinct-3 \\ \midrule
        Image-Chat & 0.739837 & 0.245980 & 0.408099 & 0.530225 \\ 
        - w. EXMODD$_{1000}$ & $+$0.0035 & $+$0.0080 & $+$0.0184 & $+$0.0023 \\ 
        - w. EXMODD$_{2000}$ & $+$0.0022 & $+$0.0072 & $+$0.0195 & $+$0.0223 \\ 
        - w. EXMODD$_{3000}$ & $+$0.0010 & $+$0.0006 & $+$0.0014 & $+$0.0011 \\ 
        - w. EXMODD$_{4000}$ & $+$0.0075 & $+$0.0053 & $+$0.0146 & $+$0.0154 \\ 
        - w. EXMODD$_{5000}$ & $+$0.0118 & $+$0.0167 & $+$0.0390 & $+$0.0441 \\ 
        - w. EXMODD$_{6000}$ & $+$0.0046 & $+$0.0093 & $+$0.0344 & $+$0.0398 \\ 
        - w. EXMODD$_{7000}$ & $+$0.0136 & $+$0.0069 & $+$0.0140 & $+$0.0135 \\ 
        - w. EXMODD$_{8000}$ & $+$0.0248 & $+$0.0048 & $+$0.0278 & $+$0.0300 \\ 
        - w. EXMODD$_{9000}$ & $+$0.0266 & $+$0.0105 & $+$0.0407 & $+$0.0422 \\ 
        - w. EXMODD$_{9989}$ & \textbf{$+$0.0300} & \textbf{$+$0.0176} & \textbf{$+$0.0605} & \textbf{$+$0.0695} \\ \bottomrule
    \end{tabular}
    \caption{\label{ALBEF Mix Dataset}
       Metrics of Minimum Loss ALBEF Model Trained on Mixed Data
    }
\end{table*}

\section{Explanation Case}
Table ~\ref{bad Case of explanation} shows some misunderstand explanation in EXMODD.
The model's primary cause of misinterpretation recognition errors during image-to-text conversion. For instance, in Case 1, the model misidentifies a \textit{goat} as a \textit{bear}. The explanation needs to portray the dialogue situation based on facts accurately, but instead, it generates an interpretation contradictory to the image truth due to misinformation. In Case 2, a \textit{hat} is mistaken for a \textit{book}, in Case 3, a \textit{cup} is wrongly identified as a \textit{cat}.


Table ~\ref{Example of explanation} shows the explanation in EXMODD and fine-tuning results on BLIP. It can be seen that BLIP mistook the sculpture as a lion in Sample 1, generating an error reply; in Sample 2, BLIP produced an illusion of \textit{water}, which could not explain the rationality of the dialogue; in Sample 3, the blip appeared logically confused, and the preceding and following sentences constituted contradictions.

\newcommand{\tabincell}[2]{\begin{tabular}{@{}#1@{}}#2\end{tabular}}
\renewcommand{\tabularxcolumn}[1]{m{#1}}
\begin{table*}
\centering
\begin{tabularx}{\textwidth}{>{\centering\arraybackslash}m{2cm} X X X}
\toprule
& \multicolumn{1}{c}{{Sample 1}} & \multicolumn{1}{c}{{Sample 2}} & \multicolumn{1}{c}{{Sample 3}} \\
\midrule
\\
\adjustbox{valign=m}{{Image}} &
\multicolumn{1}{c}{\begin{minipage}[b]{0.4\columnwidth}
    \centering
    {\includegraphics[width=95pt, height=95pt]{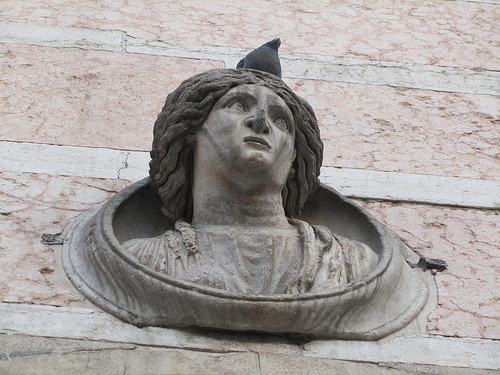}}
\end{minipage}}
&
\multicolumn{1}{c}{\begin{minipage}[b]{0.4\columnwidth}
    \centering
    {\includegraphics[width=95pt, height=95pt]{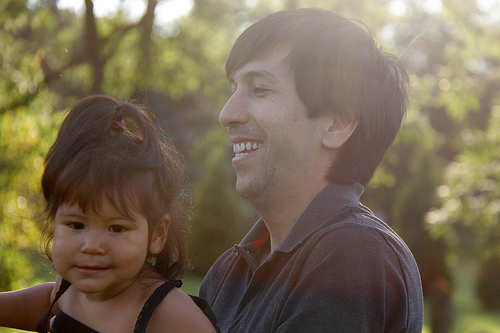}}
\end{minipage}}
&
\multicolumn{1}{c}{\begin{minipage}[b]{0.4\columnwidth}
    \centering
    {\includegraphics[width=95pt, height=95pt]{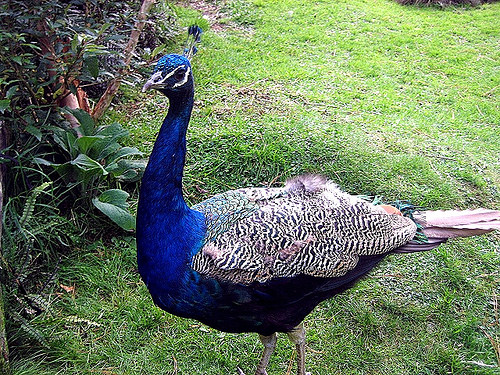}}
\end{minipage}}
\vspace{0.2cm}
\\
\midrule
\tabincell{c}{{Context}} &
I remember Medusa so well from English. &
The man should toss the girl in the air and catch her. &
Great dog you have. \\
\midrule
\tabincell{c}{{Response}} &
That statue actually depicts an elephant-headed god called Ganesha in Hindu mythology. &
That sounds like fun, but let's make sure it's safe for her first. &
Thanks, but that's actually a peacock.\\
\midrule
\tabincell{c}{{EXMODD} \\ \ {Explanation}} &
This reply is generated because the first utterance mentions a statue and its connection to English. The description of the statue in the image matches the characteristics of Ganesha, so the reply provides more information about the statue's identity and origin. &
Acknowledges the suggestion while also expressing concern for the little girl's safety. & 
The person mistook the peacock for a dog, so I created a response to correct them. \\
\midrule
\tabincell{c}{{BLIP} \\ {Explanation}} &
The first utterance mentions a statue of \textbf{a lion}, so the reply builds on that by mentioning the lion's head as a statue of a lion. &
The first utterance suggests a desire to jump into the \textbf{water}, so the reply acknowledges the desire but also suggests a safer alternative. & 
The first utterance mentions a dog, so the reply clarifies that \textbf{the dog is actually a dog}. \\
\bottomrule

\end{tabularx}
\caption{\label{Example of explanation}
Bad examples of explanation by BLIP generation. There are misidentifications, hallucinations, and logical errors in the bold part.
}
\end{table*}

\renewcommand{\tabularxcolumn}[1]{}

\renewcommand{\tabularxcolumn}[1]{m{#1}}
\begin{table*}
\centering
\begin{tabularx}{\textwidth}{>{\centering\arraybackslash}m{2cm} X X X}
\toprule
& \multicolumn{1}{c}{{Case 1}} & \multicolumn{1}{c}{{Case 2}} & \multicolumn{1}{c}{{Case 3}} \\
\midrule
\\
\adjustbox{valign=m}{{Image}} &
\multicolumn{1}{c}{\begin{minipage}[b]{0.4\columnwidth}
    \centering
    {\includegraphics[width=95pt, height=95pt]{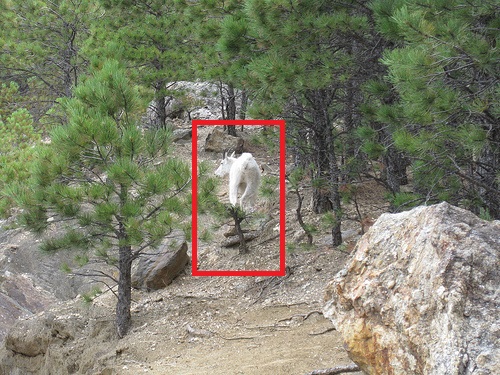}}
\end{minipage}}
&
\multicolumn{1}{c}{\begin{minipage}[b]{0.4\columnwidth}
    \centering
    {\includegraphics[width=95pt, height=95pt]{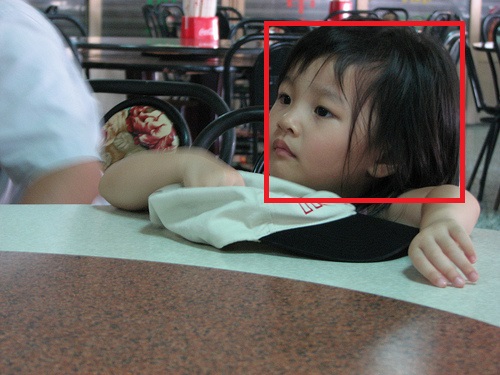}}
\end{minipage}}
&
\multicolumn{1}{c}{\begin{minipage}[b]{0.4\columnwidth}
    \centering
    {\includegraphics[width=95pt, height=95pt]{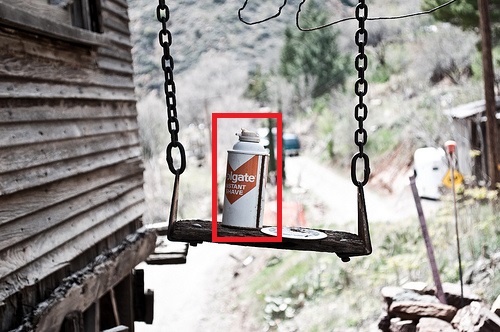}}
\end{minipage}}
\vspace{0.2cm}
\\
\midrule
\tabincell{c}{{Context}} &
Wonder if that goat has a hot date tonight? &
Little girl was making faces. &
I would never sit on a rusty swing like that one. \\
\midrule
\tabincell{c}{{Response}} &
\textcolor{red}{It's a bear}, not a goat. Please be careful, we don't want to disturb it. &
She's probably just playing around. Do you think she likes \textcolor{red}{the book}? &
Me neither, but I'd love to pet that \textcolor{red}{cute little cat} on the wooden pole.\\
\midrule
\tabincell{c}{{EXMODD} \\ {Explanation}} &
The speaker has misidentified the animal in the picture, and the reply corrects their mistake while also expressing concern for the safety of both the speaker
and \textcolor{red}{the bear}. &
The first utterance implies that the little girl was making funny facial expressions, but it is unclear why. The reply attempts to start a conversation by acknowledging the girl's behavior and then shifting the topic to ask about \textcolor{red}{the book on the table}. It is a natural and relevant response that could lead to further discussion. & 
The speaker acknowledges the first utterance and offers comments about \textcolor{red}{the cat} in the picture, which is a unique and specific detail mentioned in the scene description. This allows for further conversation and interaction between the two speakers beyond a generic response. \\
\bottomrule
\end{tabularx}
\caption{\label{bad Case of explanation}
Some bad cases of explanation in EXMODD. The red parts are inconsistent with the facts.
}
\end{table*}
\renewcommand{\tabularxcolumn}[1]{}

\end{document}